\documentclass{article}

\usepackage[nonatbib, final]{neurips_2023}

\usepackage[utf8]{inputenc} 
\usepackage[T1]{fontenc}    
\usepackage{hyperref}       
\usepackage{url}            
\usepackage{physics}
\usepackage{cite}
\usepackage{booktabs}       
\usepackage{bookmark}       
\usepackage{amsfonts}       
\usepackage{amsmath}        
\usepackage{amsthm}         
\usepackage{amssymb}
\usepackage{nicefrac}       
\usepackage{microtype}      
\usepackage{xcolor}         
\usepackage{graphicx}       
\usepackage{subcaption}     
\usepackage{svg}            
\usepackage{tikz}           
\usepackage{mathtools}
\usepackage{dsfont}

\usepackage[createShortEnv]{proof-at-the-end}
\usepackage{aligned-overset}
\usepackage{cancel}
\usepackage{pdfpages}
\usepackage{pgfplots}
\usepackage{wrapfig}  
\usepackage{multirow}
\usepackage[flushleft]{threeparttable}
\usepackage{stackengine}
\usepackage{float}
\usepackage{tabularx}

\newcolumntype{Y}{>{\centering\arraybackslash}X}

\usepackage{xspace}
\newcommand{\name}{Learn-VRF\xspace}

\title{Probabilistic Abduction for Visual Abstract Reasoning via Learning Rules in Vector-symbolic Architectures}

\author{Michael Hersche$^{1,2}$, 
Francesco di Stefano$^{1,2}$\thanks{Research conducted at IBM Research -- Zurich.}\,\,,  
Thomas Hofmann$^{2}$,\\  
\textbf{Abu Sebastian}$^{1}$,
\textbf{Abbas Rahimi}$^{1}$\thanks{Corresponding author: abr@zurich.ibm.com} \\
{
$^{1}$IBM Research -- Zurich, $^{2}$ETH Zurich
}
}

\begin{document}

\maketitle

\begin{abstract}
Abstract reasoning is a cornerstone of human intelligence, and replicating it with artificial intelligence (AI) presents an ongoing challenge. 
This study focuses on efficiently solving Raven's progressive matrices (RPM), a visual test for assessing abstract reasoning abilities, by using distributed computation and operators provided by vector-symbolic architectures (VSA). 
Instead of hard-coding the rule formulations associated with RPMs, our approach can learn the VSA rule formulations (hence the name \name) with just one pass through the training data. 
Yet, our approach, with compact parameters, remains transparent and interpretable. 
\name yields accurate predictions on I-RAVEN's in-distribution data, and exhibits strong out-of-distribution capabilities concerning unseen attribute-rule pairs, significantly outperforming pure connectionist baselines including large language models. 
Our code is available at \url{https://github.com/IBM/learn-vector-symbolic-architectures-rule-formulations}. 
\end{abstract}

\section{Introduction}
The Raven’s progressive matrices (RPM)~\cite{Raven1938} test assesses human's fluid intelligence and abstract reasoning~\cite{Carpenter1990,RavenTest2012}.
It is a non-verbal test that involves perceiving pattern continuation, elemental abstraction, and finding relations between abstract elements based on underlying rules. 
Each RPM test is an analogy problem presented as a 3×3 pictorial matrix of context panels. Every panel in the matrix is filled with several geometric objects based on a certain rule, except the last panel, which is left blank. 
The task is to complete the missing panel by picking the correct answer from a set of candidate answer panels that matches the implicit rule (see Fig.~\ref{fig:overview}).

Recently, RPM has become a widely used task for effectively testing AI capabilities in abstract reasoning, making analogies, and dealing with out-of-distribution (OOD) data~\cite{Raven_19, I-Raven,MRNet_CVPR2021,Mitchel_Survey_2021,RPM_Survey2022}.
Large language models (LLMs) have been recently shown zero-shot learning capability in solving RPMs when perceptual visual information is parsed externally and provided as symbolic inputs to LLMs~\cite{Webb2023,hu2023context}.
On the other hand, neuro-symbolic AI approaches combine subsymbolic perception with various forms of symbolic reasoning leading to state-of-the-art (SOTA) performance in visual~\cite{NS-VQA_NIPS18,NS_ConceptLearner_ICLR19,NS_MetaConcept_NIPS19,Falcon_ICLR2022}, natural language~\cite{Learn2reason_TPR}, causal~\cite{CLEVRER_ICLR2020}, mathematical~\cite{TP-transformer_2019}, and RPM analogical~\cite{PrAE_CVPR21,nesy2022_knowledge, Hersche_NMI2023} reasoning tasks. 
Specifically, for solving the RPM tests, as an upgrade of pure logical reasoning, a neuro-symbolic approach~\cite{PrAE_CVPR21} implemented a \emph{probabilistic} abduction reasoning.
Abduction refers to the process of selectively inferring facts that give the best explaination to the perceptual observations based on prior background knowledge which is represented in a symbolic form~\cite{AbductiveCognition}. 
The probabilistic abduction reasoning allowed for perceptual uncertainties in the symbol grounding which is particularly useful when connecting a trainable perception module to the reasoning module for the end-to-end training.

However, the probabilistic abduction reasoning incurs exhaustive symbolic searches.
For instance, it requires to search over the symbols describing all possible rule realizations that could govern the RPM tests to be able to abduce the probability distribution of the rules~\cite{PrAE_CVPR21}. 
A viable option to reduce such search complexity is to use distributed representations provided by vector-symbolic architectures (VSA)~\cite{VSA_03,PlateHolographic1995,PlateHolographic2003,Kanerva2009}, which is built on a rich algebra supporting well-defined operations, including multiplicative binding, unbinding, additive superposition, permutations, and similarity search that allow performing symbolic computations on top of high-dimensional distributed representations.
Consequently, a neuro-vector symbolic architecture (NVSA)~\cite{Hersche_NMI2023} was proposed to perform \emph{tractable} probabilistic abductive reasoning with distributed VSA representations and operators which offer computation-in-superposition leading to two orders of magnitude faster inference.
Nevertheless, like other neuro-symbolic approaches, NVSA had to carefully choose suitable operators and operands to formulate the individual RPM rules.
While NVSA allows the incorporation of domain knowledge into the model, it requires complete knowledge of the problem.

We propose \name, a novel probabilistic abduction reasoning approach that \emph{learns} VSA rule formulations from examples by solving a convex optimization problem. 
Besides its higher flexibility, better generality, and one-pass learning, \name maintains the advantages of neuro-symbolic approaches, including transparency and interpretability. 
\name transparently operates in the rule space by abstracting away from the surface statistical patterns.
On the I-RAVEN dataset~\cite{I-Raven}, with extracted visual attribute labels, \name achieves competitive accuracy on in-distribution data, and notably outperforms LLMs~\cite{hu2023context} and neural MLP approaches on OOD tests while having orders of magnitude lower trainable parameters.

\section{Background}

\paragraph{Vector-symbolic Architectures (VSA)}
VSA manipulates high-dimensional distributed representations by dimensionality-preserving operations, as opposed to tensor product representations which yield higher-order tensors~\cite{TP-transformer_2019,Learn2reason_TPR,Neurocompositional2022}. 
Many VSA variants exist with different representations and operators (see~\cite{VSA_Survey_Part1} for a review).
In this work, we use binary sparse block codes (SBC)~\cite{laiho2015sparse}, which induce a local blockwise structure that exhibits ideal variable binding properties~\cite{FradySDR2021}. 
In binary SBC, the binding is defined as the blockwise circular convolution ($\circledast$), and the approximate unbinding is the blockwise circular correlation ($\otimes$). 
The VSA binding has properties analogous to multiplication in the real number domain, including commutativity, associativity, and the existence of a neutral element ($\mathbf{e}$).
The cosine similarity ($\mathrm{cossim}(\cdot,\cdot)$) measures the similarity between two vectors. 

\paragraph{I-RAVEN Dataset}
The I-RAVEN dataset~\cite{I-Raven} provides RPM tests with unbiased candidate sets. 
Fig.~\ref{fig:overview} shows an I-RAVEN example test. 
Each RPM test consists of nine context panels, arranged in a 3x3 matrix, and eight candidate panels. 
The panels contain objects that are arranged according to one of seven different constellations. 
The object's attributes (color, size, shape, number, position) are governed by different underlying rules: constant, arithmetic, progression, or distribute three. 

\section{Methods}

\begin{figure}[t]
    \centering
    \includegraphics[width=\textwidth]{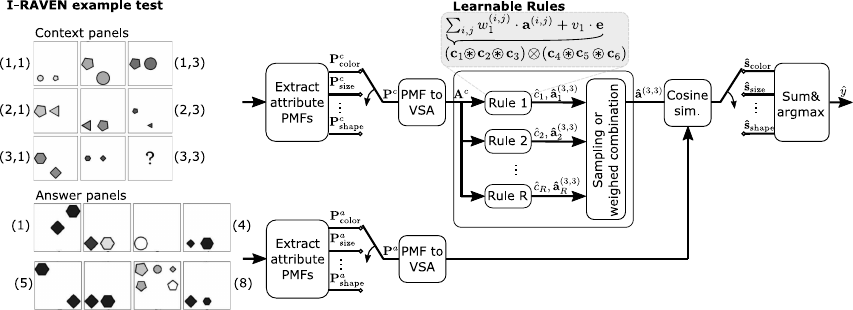}
    \caption{\name for solving Raven's progressive matrices.}
    \label{fig:overview}
\end{figure}

This section presents \name, which learns predictive VSA rule formulations based on examples, depicted in Fig.~\ref{fig:overview}.
First, the attribute values (provided by I-RAVEN metadata) are mapped to one-hot probability mass functions (PMFs), and translated to a VSA vector. 
\name learns $R$-many rules, which are shared across the attributes. 
Each rule generates a prediction with a confidence value. 
The predicted representation of the empty panel ($\mathbf{\hat{a}}^{(3,3)}$) is either based on selecting a prediction of a rule with high confidence (sampling), or a weighted combination of all the rules. 
Finally, the candidate panel with the largest sum of cosine similarities to the predicted attribute vectors is the answer.

\paragraph{Translating PMFs to VSA}

This paper focuses exclusively on the RPM's reasoning aspect; thus, we directly extract the attribute labels provided by I-RAVEN metadata, as practiced for LLMs~\cite{Webb2023,hu2023context}. 
The attributes of each object in a panel are mapped to attribute one-hot PMFs, which are further combined, yielding panel-level PMFs describing the attribute's distribution of a panel~\cite{PrAE_CVPR21}. 
Each attribute yields eight PMFs (one per context panel), represented by $\mathbf{P}^c := (\mathbf{p}^{(1,1)}, \mathbf{p}^{(1,2)}, ... \mathbf{p}^{(3,2)})$.
Next, we translate the panel-level PMFs to high-dimensional VSA representations:
\begin{align}
    \mathbf{a}^{(i,j)} = \sum_{k=1}^N \mathbf{p}^{(i,j)}[k] \cdot \mathbf{b}_k, 
\end{align}
where $N$ is the dimensionality of the panel-level PMF space.
Similarly to NVSA~\cite{Hersche_NMI2023}, we use a binary SBC dictionary $\{\mathbf{b}_k \}_{k=1}^N$, where each basis vector ($\mathbf{b}_i$) has a dimension of 1024 and four blocks. 
When combined with fractional power encoding~\cite{PlateHolographic2003}, this basis permits the representation of continuous attributes (e.g., color or size). 
The VSA representation of different context panels are represented by $\mathbf{A}^c := (\mathbf{a}^{(1,1)}, \mathbf{a}^{(1,2)}, ... \mathbf{a}^{(3,2)})$.

\begin{table}[t]
\centering
\caption{Translation of RPM rules into VSA rule formulations.}
\resizebox{0.7\textwidth}{!}{%
\begin{tabular}{ll}
\toprule
   Rule & VSA formulation \\
   \cmidrule(r){1-1}\cmidrule(r){2-2}
Constant & $\mathbf{\hat{a}}^{(3,3)} = \mathbf{a}^{(3,2)} = \mathbf{a}^{(3,1)}$\\
Arithmetic plus & $\mathbf{\hat{a}}^{(3,3)} = \mathbf{a}^{(3,1)} \circledast \mathbf{a}^{(3,2)}$ \\
Progression positive & $\mathbf{\hat{a}}^{(3,3)} = \mathbf{a}^{(3,2)} \circledast (\mathbf{a}^{(3,2)} \otimes \mathbf{a}^{(3,1)})$ \\
Distribute three & $\mathbf{\hat{a}}^{(3,3)} = (\mathbf{a}^{(1,1)} \circledast \mathbf{a}^{(1,2)}\circledast \mathbf{a}^{(1,3)}) \otimes (\mathbf{a}^{(3,1)} \circledast \mathbf{a}^{(3,2)})$\\
\bottomrule
\end{tabular}
}
\label{table:RAVEN_rules_VSA}
\end{table}

\paragraph{Learning VSA Rule Formulations (\name)}

VSA allows us to predict the representation of the empty panel with a rule-specific combination of operands (i.e., a selection of context panels) and operations (binding and unbinding). 
Table~\ref{table:RAVEN_rules_VSA} summarizes possible VSA formulations for each RPM rule. 
All the rules are represented as a sequence of binding and unbinding operations, which can be generalized to: 
\begin{align}\label{eq:fractional}
    \mathbf{\hat{a}}^{(3,3)} = (\mathbf{c}_{1} \circledast \mathbf{c}_{2} \circledast \mathbf{c}_{3}) \otimes (\mathbf{c}_{4} \circledast \mathbf{c}_{5} \circledast \mathbf{c}_{6}), 
\end{align}
where $\mathbf{c}_i$ represents a context panel ($\mathbf{a}^{(i,j)}$) or the identity ($\mathbf{e}$). 
Eq.~\eqref{eq:fractional} can be interpreted as a fractional form, akin to a sequence of multiplicative and divisive operations among real numbers.

While NVSA hard-coded the rules as in Table~\ref{table:RAVEN_rules_VSA}, we \emph{learn} the rule formulations by finding, for each element in the fractional form, a convex combination over the VSA vectors of the context panels' attributes, augmented with the neutral element: 
\begin{align}
   \mathbf{c}_k = \sum_{\text{panels } (i,j)} w_k^{(i,j)} \cdot \mathbf{a}^{(i,j)} +  v_k\cdot \mathbf{e},
\end{align}
where the following constraints apply to the weights: 
\begin{align}
   \sum_{\text{panels } (i,j)} w_k^{(i,j)} +  v_k = 1, \quad \quad 0 \leq w_k^{(i,j)} \leq 1 \, \forall i,j, \quad \quad 0 \leq v_{k} \leq 1        \, \forall k.  
\end{align}
Here, \( w_k^{(i,j)} \) and \(  v_k \) are the trainable weights of the convex combination. 
A rule in the VSA space can manifest in multiple functionally equivalent representations within the proposed form. 
As a result, the optimization problem will have multiple global minima corresponding to different algebraic representations of the same rule.

\name learns $R$-many rule formulations. 
Allowing more rules than actually embedded in I-RAVEN (i.e., $R>R^*$) yields \name to come up with functionally equivalent duplications. 
Thus, by providing an overcomplete rule space, \name can still operate and does not need to have the exact knowledge about the number of rules ($R^*$). 
In contrast, a related learnable neuro-symbolic approach~\cite{topan2021satnet} performed a grid search for the optimal number of attributes and rules in SATNet. 
For simplicity, we set the number of rules equal to the rules in I-RAVEN ($R=R^*$).

\paragraph{Confidence value computation}
Each rule ($r$) generates a VSA vector ($\mathbf{\hat{a}}^{(3,3)}_r$) accompanied by a confidence value ($\hat{c}_r$). 
The confidence is computed by deploying three formulations per rule, which predict all panels of the third column.
%
At inference, the confidence value is calculated as the sum of the cosine similarities between the predicted VSA vectors of the first and second row with their respective ground-truth vector: 
\begin{align}
    {c}_{r,\text{test}} =  \mathrm{cossim}\left(\mathbf{a}^{(1,3)},  \mathbf{\hat{a}}^{(1,3)}_r\right) + \mathrm{cossim}\left(\mathbf{a}^{(2,3)},  \mathbf{\hat{a}}^{(2,3)}_r\right).
\end{align}
During training, the ground-truth of the last row is accessible, enabling the inclusion of the cosine similarity between the predicted panel and the ground-truth answer ($\mathbf{a}_y$) into the confidence score: 
\begin{align}
    {c}_{r,\text{train}} =  \mathrm{cossim}\left(\mathbf{a}^{(1,3)},  \mathbf{\hat{a}}^{(1,3)}_r\right) + \mathrm{cossim}\left(\mathbf{a}^{(2,3)},  \mathbf{\hat{a}}^{(2,3)}_r\right) + \mathrm{cossim}\left(\mathbf{a}_y,\mathbf{\hat{a}}^{(3,3)}_{r}\right). 
\end{align}
A softmax with a temperature parameter is applied to the confidence scores, producing a rule distribution. 

\paragraph{Final estimation using sampling or weighted combination}
We distinguish between two variants to determine the final estimation of the empty panel. 
One variant is to \textit{sample} a rule ($\hat{r}$) that has a high confidence value. 
During training, we sample from the distribution of the confidence values, whereas at inference, we select the rule with the maximizing confidence value.
The estimated panel is then defined as: 
\begin{align}
    \mathbf{\hat{a}}^{(3,3)}_{\text{sample}} = \mathbf{\hat{a}}^{(3,3)}_{\hat{r}}. 
\end{align}
Alternatively, we propose to create a convex \textit{weighted combination} of all the rules using the confidence values as weights: 
\begin{align}
    \mathbf{\hat{a}}^{(3,3)}_{\text{weighted comb.}} = \sum_{r=1}^R c_r \cdot \mathbf{\hat{a}}^{(3,3)}_{r}. 
\end{align}

%
%
\paragraph{Training loss}
The model weights are updated through stochastic gradient descent (SGD), reducing the inverse cosine similarity of the selected rule as the loss: 
\begin{align}
    \mathcal{L}= 1- \mathrm{cossim}\left(\mathbf{a}^{(1,3)},  \mathbf{\hat{a}}^{(1,3)}\right) - \mathrm{cossim}\left(\mathbf{a}^{(2,3)},  \mathbf{\hat{a}}^{(2,3)}\right) - \mathrm{cossim}\left(\mathbf{a}_y,  \mathbf{\hat{a}}^{(3,3)}\right), 
\end{align}
where $\mathbf{\hat{a}}^{(i,3)}$ results either from the sampling or the weighted combination. 
The terms derived from the first and second rows refine the rule selection mechanism, whereas the term associated with the ground-truth answer adjusts the weights responsible for predicting the empty panel's VSA vector.

\section{Experimental Results}
\paragraph{I-RAVEN: in-distribution}
Table~\ref{tab:id-results} compares \name's accuracy with the SOTA deep neural nets SCL~\cite{wu2020scl}, probabilistic abductive neuro-symbolic approaches such as PrAE~\cite{PrAE_CVPR21} and NVSA~\cite{Hersche_NMI2023}, and an LLM based on GPT-3~\cite{hu2023context} on in-distribution (ID) I-RAVEN tasks. 
As an additional baseline, we trained a model that predicts the PMF for each attribute using a separate MLP.
App.~\ref{app:mlp} describes the MLP baseline and App.~\ref{app:setup} our experimental setup. 
\begin{table}[h]
\caption{ID accuracy (\%) on I-RAVEN. Selective approaches require the answer panels as model inputs, while predictive approaches compare their prediction against all possible answers. }
\label{tab:id-results}
\resizebox{\textwidth}{!}{%
\begin{threeparttable}
\begin{tabular}{lrrrcccccccc}
\toprule
                &                                                        &          &           & \multicolumn{8}{c}{I-RAVEN accuracy (\%)}                \\
                \cmidrule(r){5-12}
Method             & \begin{tabular}[c]{@{}l@{}}Predictive/\\ Selelective\end{tabular} & \#Parms & \#Epochs & Avg. & C & 2x2  & 3x3   & L-R  & U-D  & O-IC & O-IG \\
\cmidrule(r){1-4}\cmidrule(r){5-12}
{SCL~\cite{wu2020scl}} & Selective & 961\,k & 200 &  {84.3} &   {99.9} &  {68.9} &  {43.0} &  {98.5} &  {99.1} &  {97.7} &  {82.6} \\
GPT-3~\cite{hu2023context} & Selective & 175\,B & n/a & 86.5 & 86.4 & 83.2 & 81.8 & 83.4 & 84.6 & 92.8 & 93.0 \\
{PrAE\tnote{1}~\cite{PrAE_CVPR21}}   & Predictive & - &  -            &  71.1 &  83.8 & 82.9 &47.4  & 94.8 & 94.8 & 56.6  & 37.4 \\
NVSA\tnote{1}~\cite{Hersche_NMI2023}    &Predictive &   -             &  -         & 88.1 & 99.8   & 96.2 & 54.3  & 100  & 99.9 & 99.6 & 67.1 \\
\cmidrule(r){1-4}\cmidrule(r){5-12}
MLP baseline  & Predictive & 300\,k     & 50        & 87.1 & 97.6   & 87.1 & 61.8  & 99.4 & 99.4 & 98.7 & 65.6 \\
\name (sampling)     & Predictive & 5\,k       & 50         & 81.3 & 98.7   & 62.9 & 48.4& 98.7 & 99.9 & 99.9 & 60.8 \\
\name (weighted comb.)     & Predictive & 5\,k       & 50         & 84.1 & 99.1 & 73.5  & 50.9 & 99.9 & 99.9 & 100  & 65.2 \\
\name (sampling)    & Predictive & 5\,k       & 1         & 80.0 & 99.1   & 52.3 & 51.1 & 98.7 & 99.9 & 99.9 & 59.5 \\
\name (weighted comb.)     & Predictive & 5\,k       & 1         & 81.0 & 99.1 & 53.1  & 50.9 & 99.9 & 100 & 100  & 63.7 \\
\bottomrule
\end{tabular}
  \begin{tablenotes}
    \item[1] No trainable parameters in the hard-coded reasoning; hence, learnable parameters and epochs are zero for reasoning module.
  \end{tablenotes}
\end{threeparttable}
}
\end{table}

\name achieves an average accuracy of 81.3\% when using a sampling-based rule selection and 84.1\% with the weighted rule combination, while only requiring 5\,k trainable parameters.
The superior performance of the weighted combination over the sampling approach may stem from its uniform update mechanism, which adjusts all the rules (including the ``correct'' rule) instead of only the sampled one. 
\name remains accurate when training it with a single pass over the training data, achieving 80.0\% and 81.0\% using the sampling and weighted combination, respectively. 
\name generates accurate predictions in constellations with no position attribute (i.e., center, left-right, up-down, out-in center). 
In these constellations, it outperforms GPT-3 and PrAE, and is on par with NVSA and SCL. 
Instead, the 2x2 grid, 3x3 grid, and out-in grid require the position attribute that employs logical rules on its bit-code vector, which the VSA rule formulations cannot support (see App.~\ref{app:limitations}). 
While NVSA circumvented this limitation by treating some rules of the position attribute in the original PMF domain, \name uniformly implements all rules in the VSA space, at some degree of accuracy degradation.  
The MLP baseline can improve the accuracy in the position-related constellations, but it requires 60$\times$ more parameters (300\,k vs. 5\,k) while also sacrificing transparency and interpretability.


\paragraph{I-RAVEN: out-of-distribution} 
We test GPT-3, the MLP baseline, and the \name on unseen rule-attribute pairs in the center constellation, validating their OOD generalization capabilities (see App.~\ref{app:setup}).
In this OOD task, \name has been trained for 50 epochs.
As shown in Table~\ref{tab:results-ruleattribute-generalization}, the MLP baseline faces challenges in OOD even though it was accurate on ID. 
%
%
Moreover, the GPT-3 shows a low accuracy, particularly on the arithmetic rule, despite the potential for data contamination~\cite{saparov2023language}, and our additional efforts to perform in-context learning (see App.~\ref{app:llm}). 
In contrast, \name generalizes to OOD, achieving almost perfect accuracy on most rule-attribute pairs thanks to \name's rule sharing, which is enabled by the uniform, expressive PMF representation in the VSA space using fractional power encoding. 

\begin{table}[h]
\caption{OOD accuracy (\%) on unseen rule-attribute pairs on I-RAVEN in the center constellation. 
} 
\label{tab:results-ruleattribute-generalization}
\resizebox{\textwidth}{!}{
\begin{threeparttable}
\begin{tabular}{llllllllllll}
\toprule
        & \multicolumn{3}{c}{Type} & \multicolumn{4}{c}{Size}  & \multicolumn{4}{c}{Color} \\
        \cmidrule(r){2-4} \cmidrule(r){5-8} \cmidrule(r){9-12}
        & Const.      & Progr.      & Dist.3      & Const.      & Progr.      & Dist.3 & Arith.    & Const.     & Progr.      & Dist.3 & Arith.  \\
        \cmidrule(r){1-1} \cmidrule(r){2-4} \cmidrule(r){5-8} \cmidrule(r){9-12}
GPT-3 zero-shot & 88.5 & 86.0 & 88.6  & 93.6 &93.2 & 92.6 & 71.6 & 94.2 & 94.7 & 94.3 & 65.8\\
MLP baseline & 14.8 & 14.9 & 30.2 & 22.8 & 75.0 & 47.0 & 46.6 & 56.3 & 60.5 & 44.4 & 48.9\\
\name (weighted comb.) & 100 & 100 & 99.7 & 100 & 100 & 99.8 & 99.8 & 100 & 98.8 & 100 & 100 \\
\bottomrule
\end{tabular}
\end{threeparttable}
}
\end{table}

\paragraph{Interpretation of learned rule formulations} \name provides transparency and interpretability into the learned rules.
We can derive the most dominant formula by selecting the VSA vector that exhibited the highest weight for each term within the \name model's internal representations. 
Table~\ref{tab:learned_rules} shows the learned rule formulations for the center constellation. 
Analyzing these rules and performing straightforward algebraic simplifications, we discern the equivalences shown in the second column. 
App.~\ref{app:rules} provides more details on the interpretation of the learned rule formulations.

\begin{table}[H]
\caption{Rule formulations learned by \name with the respective rule interpretation.}
\label{tab:learned_rules}
\centering
\resizebox{.8\textwidth}{!}{
\begin{tabular}{ll}
\toprule
Extracted Learned Rules & Corresponding RPM Rule \\
\hline
$\mathbf{\hat{a}}^{(3,3)} = (\mathbf{a}^{(3,2)} \circledast \mathbf{a}^{(1,1)} \circledast \mathbf{a}^{(3,1)}) \otimes (\mathbf{e} \circledast \mathbf{a}^{(1,2)} \circledast \mathbf{a}^{(3,2)})$  & dist. three right \& constant \\
 $\mathbf{\hat{a}}^{(3,3)} = (\mathbf{a}^{(3,2)} \circledast \mathbf{a}^{(1,2)} \circledast \mathbf{a}^{(2,3)}) \otimes (\mathbf{a}^{(2,3)} \circledast \mathbf{e} \circledast \mathbf{a}^{(1,1)})$  & dist. three left \& constant \\
$\mathbf{\hat{a}}^{(3,3)} = (\mathbf{a}^{(3,2)} \circledast \mathbf{a}^{(3,1)} \circledast \mathbf{a}^{(3,1)}) \otimes (\mathbf{e} \circledast \mathbf{a}^{(3,1)} \circledast \mathbf{e})$  & arithmetic plus\\
$\mathbf{\hat{a}}^{(3,3)} = (\mathbf{a}^{(3,2)} \circledast \mathbf{a}^{(3,2)} \circledast \mathbf{a}^{(3,1)}) \otimes (\mathbf{a}^{(3,2)} \circledast \mathbf{a}^{(3,2)} \circledast \mathbf{a}^{(3,2)})$  & arithmetic minus\\
$\mathbf{\hat{a}}^{(3,3)} = (\mathbf{a}^{(1,3)} \circledast \mathbf{a}^{(3,1)} \circledast \mathbf{a}^{(3,2)}) \otimes (\mathbf{e} \circledast \mathbf{a}^{(3,2)} \circledast \mathbf{a}^{(1,1)})$  & progression \\
\bottomrule
\end{tabular}
}
\end{table}

\section{Conclusion}
We proposed \name, a novel VSA-based approach for probabilistic abduction which can learn rule formulations. 
The model's structure inherently provides transparency: by identifying the index with the highest probability mass for each term in the weight distribution, we could reconstruct the rule formula chosen by the \name. 
Moreover, \name features a low parameter count, single-pass learning, and OOD generalization with respect to unseen attribute-rule combinations. 

\section*{Acknowledgement}
This work is supported by the Swiss National Science foundation (SNF), grant 200800.



\newpage
\appendix
\setcounter{figure}{0}
\renewcommand{\thefigure}{A\arabic{figure}}
\setcounter{table}{0}
\renewcommand{\thetable}{A\arabic{table}}

\section{MLP Baseline}\label{app:mlp}

\begin{figure}[t]
    \centering
    \includegraphics[width=.75\textwidth]{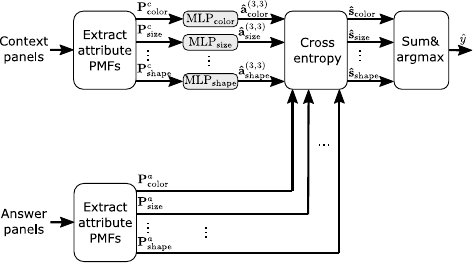}
    \caption{MLP baseline which uses one MLP per each attribute.}
    \label{fig:baseline}
\end{figure}

This baseline, illustrated in Fig.~\ref{fig:baseline}, is built upon the use of $A$ instances of MLPs, where $A$ denotes the total number of attributes. 
The MLP directly models the transformations of the attribute PMFs. 
For every attribute, there is a dedicated MLP, ensuring that each attribute is modeled independently.
Since the PMFs of the attributes have different dimensionality (i.e., different number of possible values), the MLPs cannot be shared across the attributes, while it is possible with our \name thanks to the use of VSA-based fractional power encoding that allows to map any PMF vectors with  arbitrary sizes to the fixed-width, yet expressive, distributed representation.
Thanks to this \emph{uniform} interface, \name can be plugged-and-played from one attribute to another.

The first step of MLP baseline involves processing the derived attribute values. These values are converted into one-hot PMFs. Once translated to the PMF format, these representations are processed by a 3-layer MLP with 1024 hidden dimensions in each layer. The output from this MLP is another PMF spanning the attribute values. For each attribute, its associated PMF vector is compared against the PMF vectors of all candidate panels using the cross entropy. 
For every attribute, we obtain eight scores corresponding to the eight candidate panels. Lastly, we compute the average for each attribute, resulting in eight overall candidate scores. The candidate panel with the highest score is selected.

This baseline does not make any distinction between rule selection and execution. 
Unlike \name, MLPs can potentially capture non-linear relationships, such as the logical rules present in the position attribute in the 2x2 grid, 3x3 grid, and the out-in grid. 

\section{Experimental Setup}\label{app:setup}

\subsection*{Training setup}
In our experimental setup with the MLP baselines, we used three hidden layers, each containing 1024 neurons. Layer normalization was applied between consecutive layers. We set the learning rate to 1e-4, used a batchsize of 32, and trained the model for 50 epochs. The model with the lowest validation loss from these epochs was chosen for further evaluation on the test dataset.

In \name, we used VSA vectors of 1024 dimensions with 4 blocks. Moreover, we used a learning rate of 0.1 and batchsize of 4. 
We set the number of learnable rules ($R$) equal to the number of rules in I-RAVEN.
Finally, we used a softmax temperature of 0.01, favoring exploitation over exploration, allowing us to converge quickly.

\subsection*{OOD test setup}
In the out-of-distribution (OOD) experiments, we evaluate whether the model can solve an unseen target attribute-rule pair (e.g., the constant rule on the type attribute) when it has been trained on the examples containing all of the attribute-rule pairs except the specific target one (e.g., the constant rule on size and color, the progression rule on all attributes, and the distribute rule on all attributes). 
We generate a new training and validation set containing all examples except those with the target attribute-rule pair and a test set containing examples exclusively with the target attribute-rule pair. 

In these experiments, we focus on the center single constellation because it encompasses a single panel. 
In contrast, configurations such as up-down, left-right, and in-out entail two center single constellations, introducing ambiguity in applying attribute-rule filtering.
The datasets are generated by filtering the existing splits I-RAVEN. 
The training sets encompass between 2622 and 3437 samples, the validation sets range from 841 to 1160 samples, while the test sets span from 803 to 1117 samples.

\section{Interpretation of Learned Rule Formulation in \name}
\label{app:rules}
Within \name, a singular rule in the VSA space can be depicted through various synonymous representations. 
As an example, let us examine the arithmetic plus rule for the last row:
\begin{align}
    \mathbf{\hat{a}}^{(3,3)} = \mathbf{a}^{(3,1)} \circledast \mathbf{a}^{(3,2)}. 
\end{align}
This rule can be reformulated in the VSA fractional form in multiple manners, an example subset of all the possible algebraic forms of this rule is the following one:
\begin{align}
    \mathbf{\hat{a}}^{(3,3)} &= (\mathbf{a}^{(3,1)} \circledast \mathbf{a}^{(3,2)} \circledast \mathbf{e}) \otimes (\mathbf{e} \circledast \mathbf{e} \circledast \mathbf{e}) \\
    \mathbf{\hat{a}}^{(3,3)} &= (\mathbf{a}^{(3,2)} \circledast \mathbf{a}^{(3,1)} \circledast \mathbf{e}) \otimes (\mathbf{e} \circledast \mathbf{e} \circledast \mathbf{e}) \\
    \mathbf{\hat{a}}^{(3,3)} &= (\mathbf{a}^{(3,1)} \circledast \mathbf{a}^{(3,2)} \circledast \mathbf{a}^{(3,2)}) \otimes (\mathbf{a}^{(3,2)} \circledast \mathbf{e} \circledast \mathbf{e})
\end{align}
When employing a regression loss function to the resultant VSA vector from the \name and to its related target VSA vector, the loss curve's global minima align with diverse algebraic interpretations of the identical rule. 

\section{Model Limitations}\label{app:limitations}

The primary constraint of the \name model lies in its expressiveness. To understand this, consider that in the VSA space, the sum of two attribute values is represented by the binding of their respective VSA vectors, while the difference is depicted by the unbinding operation. Given the VSA fractional form underpinning the \name:

\begin{align}
    \mathbf{\hat{a}}^{(3,3)}=(\mathbf{a}_{1} \circledast \mathbf{a}_{2} \circledast \mathbf{a}_{3}) \otimes (\mathbf{a}_{4} \circledast \mathbf{a}_{5} \circledast \mathbf{a}_{6})
\end{align}

In the numeric attribute space, this translates to:

\begin{align}
    {\hat{a}}^{(3,3)}=(a_{1} + a_{2} + a_{3}) - (a_{4} + a_{5} + a_{6})
\end{align}

This implies that the model is limited to representing sums and differences of attributes, precluding the possibility of modeling linear combinations. This limitation becomes evident in complex constellations like the 2x2 grid, the 3x3 grid, or the out-in grid, where the position attribute employs logical rules on its bit-code vector.

Table~\ref{tab:number-position} and Table~\ref{tab:other-attributes} show the number of rule occurrences of the number/position and type/size/color attributes, respectively. 
On the position attribute, the logical rules (progression and arithmetic) make up 23.9\%, 26.7\%, and 26.6\% on the 2x2 grid, 3x3 grid, and out-in grid constellation, respectively. 
Since \name cannot represent these logical rules, an upper bound for the accuracy can be determined by the share of non-logical rules (i.e, 76.1\%, 73.3\%, and 73.4\%) 
Indeed, \name (weighted combination, 50 epochs) reaches almost the upper bound on the 2x2 grid (73.5\% vs. 76.1\%), whereas the performances on the 3x3 grid (50.9\% vs. 73.3\%) and out-in grid (65.2\% vs. 73.4\%) are slightly behind.

\begin{table}[htbp]
\caption{Number of rule occurrences on the position and number attribute in the I-RAVEN test set (2000 samples). The progression and arithmetic rule on position attribute are logical rules, e.g., I-RAVEN implements the arithmetic plus rule as logical OR.}
\label{tab:number-position}
\resizebox{\textwidth}{!}{%
\centering
\begin{tabular}{lrrrrrrrrrr}
\toprule
                  & Number/Position &  \multicolumn{3}{c}{Number} & \multicolumn{3}{c}{Position} & \multicolumn{1}{c}{\multirow{2}{*}{\begin{tabular}[c]{@{}c@{}}Share of\\ non-logical\\ rules\end{tabular}}} & \multicolumn{1}{c}{\multirow{2}{*}{\begin{tabular}[c]{@{}c@{}}Acc. \name\\ (weighted comb.,\\ 50 epochs)\end{tabular}}}  \\
\cmidrule(r){2-2}\cmidrule(r){3-5}\cmidrule(r){6-8}
\textbf{} &
  \multicolumn{1}{r}{Const.} &
  \multicolumn{1}{l}{Progr.} &
  \multicolumn{1}{l}{Arith.} &
  \multicolumn{1}{l}{Dist.3} &
  \multicolumn{1}{l}{\begin{tabular}[c]{@{}c@{}}Progr.\\ (logical)\end{tabular}} &
  \multicolumn{1}{l}{\begin{tabular}[c]{@{}c@{}}Arith.\\ (logical)\end{tabular}} &
  \multicolumn{1}{l}{Dist.3} &  
  \\
\cmidrule(r){1-1}\cmidrule(r){2-2}\cmidrule(r){3-5}\cmidrule(r){6-8}\cmidrule(r){9-9}\cmidrule(r){10-10}
{2x2 grid} & 320  & 178 & 335 & 326 & 163 & 316 & 362 & 76.1\% &  73.5\%   \\
{3x3 grid} & 267  & 304 & 312 & 269 & 247 & 287 & 314 & 73.3\%  & 50.9\% \\
O-IG (inner grid)             & 325 & 158 & 321 & 342 & 175 & 357 & 322 & 73.4\% & 65.2\% \\
\bottomrule
\end{tabular}%
}
\end{table}

\begin{table}[htbp]
\caption{Number of rule occurrences on the type, size, and color attribute in the I-RAVEN test set (2000 samples). The sum of occurrences per attribute is 2000 for single constellations (center, 2x2 grid, 3x3 grid), and 4000 for dual constellations (L-R, U-D, O-IC, O-IG).  }
\label{tab:other-attributes}
\resizebox{\textwidth}{!}{%
\begin{tabular}{lrrrrrrrrrrrr}
\toprule
                  & \multicolumn{4}{c}{Type} & \multicolumn{4}{c}{Size} & \multicolumn{4}{c}{Color} \\
\cmidrule(r){1-1}\cmidrule(r){2-5}\cmidrule(r){6-9}\cmidrule(r){10-13}
\textbf{} &
  \multicolumn{1}{l}{Const.} &
  \multicolumn{1}{l}{Progr.} &
  \multicolumn{1}{l}{Arith.} &
  \multicolumn{1}{l}{Dist.3} &
  \multicolumn{1}{l}{Const.} &
  \multicolumn{1}{l}{Progr.} &
  \multicolumn{1}{l}{Arith.} &
  \multicolumn{1}{l}{Dist.3} &
  \multicolumn{1}{l}{Const.} &
  \multicolumn{1}{l}{Progr.} &
  \multicolumn{1}{l}{Arith.} &
  \multicolumn{1}{l}{Dist.3} \\
\cmidrule(r){1-1}\cmidrule(r){2-5}\cmidrule(r){6-9}\cmidrule(r){10-13}
{Center}   & 660   & 665   & 0 & 675  & 514  & 485  & 500 & 501  & 520   & 494  & 474 & 512  \\
{2x2 grid} & 625   & 676   & 0 & 669  & 515  & 503  & 492 & 490  & 507   & 495  & 519 & 479  \\
{3x3 grid} & 686   & 657   & 0 & 657  & 466  & 546  & 464 & 524  & 494   & 476  & 506 & 524  \\
{L-R}      & 1294  & 1306  & 0 & 1400 & 1031 & 983  & 972 & 1014 & 1006  & 1019 & 943 & 1032 \\
{U-D}      & 1332  & 1364  & 0 & 1304 & 989  & 1030 & 994 & 987  & 983   & 1015 & 983 & 1019 \\
{O-IC}     & 1278  & 1359  & 0 & 1363 & 1286 & 934  & 503 & 1277 & 2490  & 493  & 501 & 516  \\
O-IG              & 1334  & 1646  & 0 & 1320 & 1383 & 709  & 551 & 1357 & 2493  & 534  & 516 & 457 \\
\bottomrule
\end{tabular}%
}
\end{table}

\section{LLM Experiments}\label{app:llm} 
This appendix describes our additional experiments testing the OOD generalization of GPT-3 on I-RAVEN. 
Our implementation is based on~\cite{brown2020gpt3}, which uses a \texttt{text-davinci-002} model.

The attribute labels are transformed into prompts using two different approaches. 
In the entangled setting (aka entity attribute naming), the different attributes of an object are represented in one entity (e.g., packed in brackets), yielding one prompt. 
Conversely, the attributes are fully decomposed into one prompt per attribute in the disentangled setting (aka attribute decomposition).  
Both approaches perform selective classification, i.e., prompting the LLM with every answer panel in the context matrix and selecting the candidate panel as the answer that yielded the highest sum of log-probabilities in the answer part. 
On ID data, explicit disentanglement improves LLM's reasoning: in the center constellation, the GPT-3 achieves an accuracy of 86.4\% with disentanglement and 80.8\% without disentanglement~\cite{brown2020gpt3}. 

We tested the GPT-3 on our OOD test set using the disentangled setting. 
Note that we do not have complete knowledge about the data used to train GPT-3. Hence, data contamination is possible, i.e., the model could have been trained on RPMs. As a result, there is no certainty that this test is truly OOD; creating new synthetic datasets would be needed to have full control over the OOD tests as it has been done in a synthetic question-answering dataset~\cite{saparov2023language}. 
Table~\ref{tab:results-ruleattribute-generalization} shows that GPT-3 mainly faces challenges with the arithmetic rule, achieving 71.6\% on the size attribute and 65.8\% on color. 

\begin{figure}
    \centering
    \includegraphics[width=0.9\textwidth]{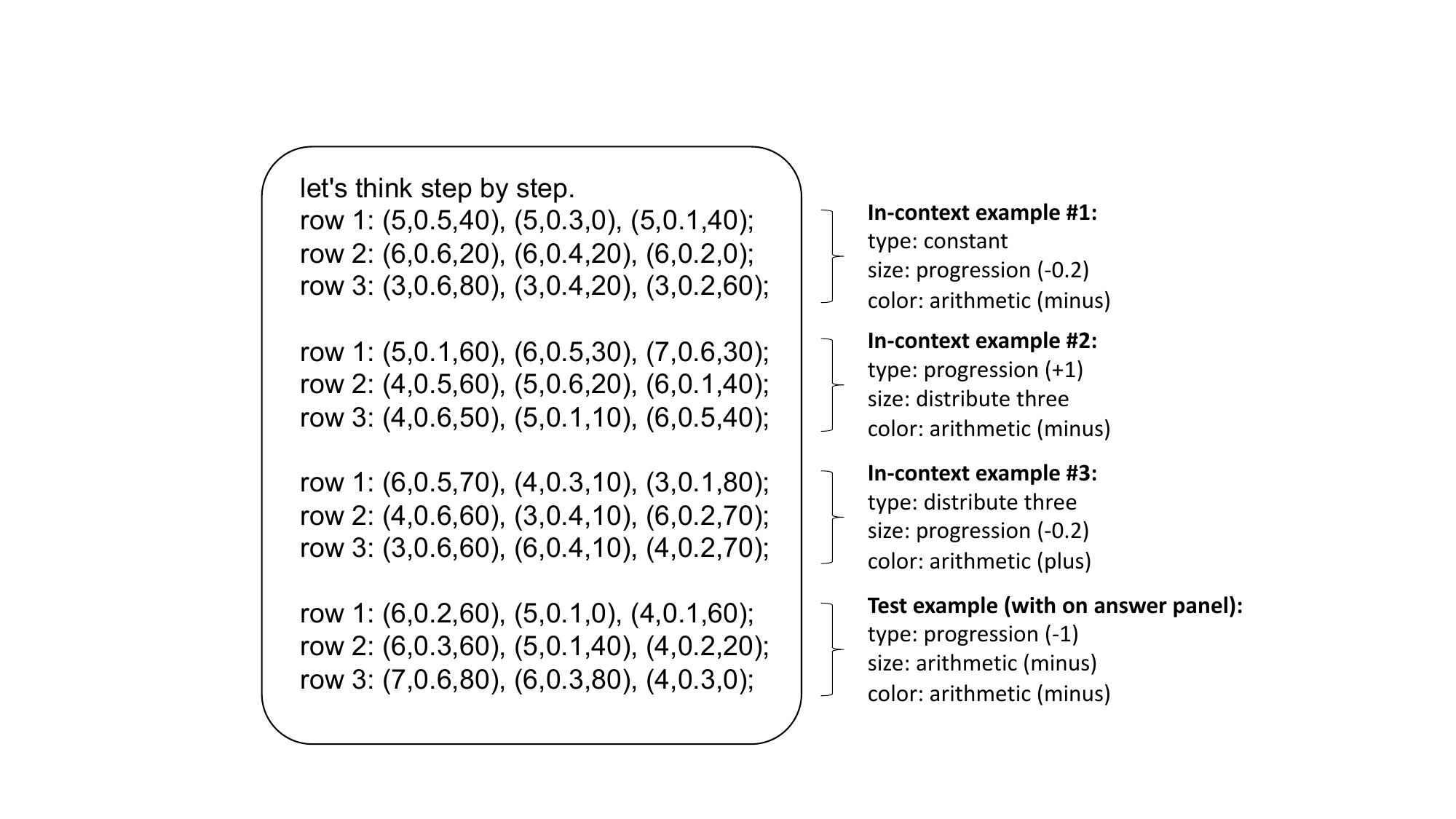}
    \caption{Prompt for GPT-3 experiment with three in-context examples in entangled setting. The single object in the center constellation is encoded into (type, size color). The OOD test focuses on the arithmetic rule on the attribute size. In-context examples show examples of arithmetic rule on attribute color. In this prompt, the empty panel in the text example is filled with the first candidate panel, which is not the correct one. }
    \label{fig:in-context}
\end{figure}

As an additional experiment, we tested the effect of in-context learning on the arithmetic rules. 
We randomly selected in-context examples that contained the arithmetic rule on an attribute different from the one under test, e.g., for the OOD test with the arithmetic on the attribute size, we showed additional examples where arithmetic was applied on color (but not size). 
As no in-context examples could be shown in the disentangled configuration, we switched to entangled setting.  
Fig.~\ref{fig:in-context} shows an example. 
To reduce the number of prompts, we only tested on a subset of 50 samples, which still gives an indication of the trends. 
The results shown in Table~\ref{tab:ood-arithmetic} indicate that in-context learning is rather detrimental in this application. 
Further efforts such as prompt-tuning (e.g., the in-context selection and arrangement) may improve the LLM's in-context learning performance~\cite{qiu2022semanticparsing}. 

\begin{table}[h]
\caption{Classification accuracy (\%) on a subset (50 samples) of I-RAVEN OOD test set  using GPT-3 (entangled setting) using a different number of in-context examples.}
\label{tab:ood-arithmetic}
\centering
\begin{tabular}{lcc}
\toprule
\#In-context samples & Size & Color \\
\cmidrule(r){1-1} \cmidrule(r){2-3}
0                  & 70.0 & 70.0  \\
3                  & 70.0 & 52.0  \\
5                  & 60.0 & 48.0  \\
10                 & 58.0 & 46.0 \\
\bottomrule
\end{tabular}
\end{table}


\end{document}